\documentclass[sigconf]{acmart}

\usepackage{multirow}
\usepackage{algorithm,algorithmic}
\usepackage{subfigure}
\usepackage{float}
\usepackage{stfloats}
\usepackage{graphicx}
\usepackage{autobreak}
\usepackage[bottom]{footmisc}
\AtBeginDocument{%
  \providecommand\BibTeX{{%
    \normalfont B\kern-0.5em{\scshape i\kern-0.25em b}\kern-0.8em\TeX}}}

\copyrightyear{2022}
\acmYear{2022}
\setcopyright{acmlicensed}
\acmConference[KDD '22] {Proceedings of the 28th ACM SIGKDD Conference on Knowledge Discovery and Data Mining}{August 14--18, 2022}{Washington, DC, USA.}
\acmBooktitle{Proceedings of the 28th ACM SIGKDD Conference on Knowledge Discovery and Data Mining (KDD '22), August 14--18, 2022, Washington, DC, USA}
\acmPrice{15.00}
\acmISBN{978-1-4503-9385-0/22/08}
\acmDOI{10.1145/3534678.3539093}

\begin{document}

\title{Uncertainty Quantification of Sparse Travel Demand Prediction with Spatial-Temporal Graph Neural Networks}


\author{Dingyi Zhuang}
\affiliation{
\institution{Massachusetts Institute of Technology}
\city{Cambridge}
\state{Massachusetts}
\country{USA}
}
\email{dingyi@mit.edu}

\author{Shenhao Wang}\authornote{Shenhao Wang is the corresponding author.}
\affiliation{
\institution{Massachusetts Institute of Technology}
\city{Cambridge}
\state{Massachusetts}
\country{USA}
}

\affiliation{
\institution{University of Florida}
\city{Gainesville}
\state{Florida}
\country{USA}
}
\email{shenhao@mit.edu}
\email{shenhaowang@ufl.edu}

\author{Haris N. Koutsopoulos}
\affiliation{\institution{Northeastern University}
\city{Boston}
\state{Massachusetts}
\country{USA}
}
\email{h.koutsopoulos@northeastern.edu}

\author{Jinhua Zhao}
\affiliation{
\institution{Massachusetts Institute of Technology}
\city{Cambridge}
\state{Massachusetts}
\country{USA}
}
\email{jinhua@mit.edu}



\renewcommand{\shortauthors}{Zhuang, Dingyi, et al.}

\begin{abstract}
Origin-Destination (O-D) travel demand prediction is a fundamental challenge in transportation. Recently, spatial-temporal deep learning models demonstrate the tremendous potential to enhance prediction accuracy. However, few studies tackled the uncertainty and sparsity issues in fine-grained O-D matrices. This presents a serious problem, because a vast number of zeros deviate from the Gaussian assumption underlying the deterministic deep learning models. To address this issue, we design a Spatial-Temporal Zero-Inflated Negative Binomial Graph Neural Network (STZINB-GNN) to quantify the uncertainty of the sparse travel demand. It analyzes spatial and temporal correlations using diffusion and temporal convolution networks, which are then fused to parameterize the probabilistic distributions of travel demand. The STZINB-GNN is examined using two real-world datasets with various spatial and temporal resolutions. The results demonstrate the superiority of STZINB-GNN over benchmark models, especially under high spatial-temporal resolutions, because of its high accuracy, tight confidence intervals, and interpretable parameters. The sparsity parameter of the STZINB-GNN has physical interpretation for various transportation applications.
\end{abstract}





\maketitle

\section{Introduction}


It has attracted a lot of attention to predict the Origin-Destination (O-D) matrices of travel demand \citep{ke2021predicting,xiong2020dynamic,jiang2022deep}. This task is challenging because the characteristics of the demand data vary with mobility services. For example, the O-D matrices for the subway stations are generally dense and probably satisfy the continuous Gaussian distribution, which implicitly underlies deterministic deep learning models. But for ride-hailing or bike-sharing services, the O-D zones could be much more granular, which lead to sparse and discrete entries in the O-D matrices. This sparsity issue becomes even more severe in the O-D matrices with high spatial-temporal resolution, because the initially dense O-D matrices could become much more diluted. As the mobility companies are increasingly adopting real-time interventions, the accurate prediction of the sparse and discrete O-D matrices at a high spatial-temporal resolution could significantly improve the quality of mobility services. 

Although the prediction of dense and low-resolution O-D matrices has been extensively studied using deep learning (DL) models, few studies addressed the sparsity issue in the high-resolution travel demand data. The continuous data entries in the dense O-D matrices could follow Gaussian distributions \citep{ke2017short,sun2020predicting,zhuang2020compound,ke2021predicting,zheng2021equality, liu2022universal}. However, a large number of zero entries in a sparse O-D matrix evidently deviate from the Gaussian assumption. When the O-D matrix is sparse and dispersed with integer values, discrete distributions like the negative binomial distribution, would be more appropriate. Moreover, an enormous number of zeros could naturally emerge when the data set has a high spatial-temporal resolution. These zeros are important for transportation management, because they indicate areas with particularly low demand. Therefore, a successful prediction model should capture explicitly the zeros in the sparse matrix and quantify their uncertainty, thus guiding the service allocation and management decisions.


To address sparsity, we propose a Spatial-Temporal Zero-Inflated Negative Binomial Graph Neural Network (STZINB-GNN) to quantify uncertainty and enhance prediction performance. We utilize zero-inflated negative binomial (ZINB) distributions to capture the enormous number of zeros in sparse O-D matrices, and the negative binomial (NB) distribution for each non-zero entry. Different from the variational autoencoder models, we design the spatial-temporal embedding with an additional parameter $\pi$ to learn the likelihood of the inputs being zero. Our model utilizes the representation power of spatial-temporal graph neural networks to fit the parameters of probabilistic distributions. We compare a variety of probabilistic layers to assess the effectiveness of $\pi$ in capturing sparsity. Empirically, we demonstrate the superiority of our model in the data set with fine-grained spatial-temporal resolution. Our main contributions include:
\begin{enumerate}
    \item We propose the STZINB-GNN to quantify the spatial-temporal uncertainty of O-D travel demand using a parameter $\pi$ to learn data sparsity
    
    \item The parameters of the probabilistic GNNs successfully quantify the sparse and discrete uncertainty particularly in high-resolution data sets
    
    \item We demonstrate that the STZINB-GNN outperforms other models by using two real-world travel demand datasets with various spatial-temporal resolutions
\end{enumerate}

The paper is organized as follows. Section \ref{sec:literature} summarizes recent studies related to DL models for travel demand prediction, sparse data modeling, and uncertainty quantification. Section \ref{sec:method} defines the research question and develops the model. Section \ref{sec:experiment} introduces the dataset used for the case study, the evaluation metrics, and the experimental results. Section \ref{sec:conclusion} concludes the paper and discuss future research.

\section{Literature Review}
\label{sec:literature}

\subsection{Travel demand prediction with spatial-temporal deep learning}
Travel demand prediction is a fundamental task in transportation applications. Based on the spatial division of the area of interest into zones, travel demand prediction is usually associated with the prediction of flow from the origin zone to the destination zone pairs, in the form of O-D matrices \citep{ke2021predicting}. The main challenges for O-D matrix prediction relate to time series prediction and spatial correlation detection. Recently these challenges have been addressed using spatial-temporal deep learning techniques. Convolution Neural Network (CNN) based techniques are applied to extract the spatial patterns because O-D matrices are usually modelled on urban grids \citep{Zhang2017DeepPrediction,liu2019contextualized,yao2018deep}. Noticeably, the CNN and its variants discover spatial patterns in Euclidean space \citep{wu2020comprehensive}. The O-D matrices naturally possess the graph structure where origin or destination regions are usually regarded as the nodes and the O-D pairs are the edges. Recent work also applied graph neural networks (GNNs) on travel demand estimation to capture non-Euclidean correlations \citep{Geng2019SpatiotemporalForecasting,xiong2020dynamic}. On the other hand, \citet{ke2021predicting} regarded O-D pairs, instead of origin and destination regions, as nodes and applied multi-graph convolutional network to predict ride-sourcing demand. Other studies used deep learning to predict the individual travel behavior by focusing on interpretation and architectural design \citep{Wang2020_interpretation,Wang2020_architecture}. The power of deep learning arises from its representation learning capacity and its new statistical foundation \citep{Wang2021_stat_learning}. However, both CNN and GNN-based approaches treat the O-D matrix entries as continuous and only focus on coarse temporal resolutions, like 60 minutes. Very few studies discussed the challenge to analyze the sparse O-D matrices with high spatial-temporal resolution.

Sparse travel demand data are different from the missing data. Zeros in sparse data mean no trips, but the sensors work properly. Missing data, on the other hand, are unobservable, potentially due to sensor malfunction. \citet{wang2021low} applied the spatial-temporal Hankelization and tensor factorization to estimate traffic states using only 17\% of the matrix entries. However, tensor factorization is a transductive method, which needs recalculation when new data come in. It is not suitable for short-term prediction. To differentiate our contributions from previous work, we will emphasize our model's spatial-temporal interpretability and uncertainty quantification in travel demand prediction.

\subsection{Uncertainty of sparse travel demand prediction}
Besides the average value, it is widely acknowledged that models should also predict uncertainty \cite{Zhao2002TheAnalysis,Rasouli2012UncertaintyAgenda}. The deterministic models that dominate the majority of the research implicitly assume homoskedasticity (i.e. common variance), which significantly simplifies the variance structure \citep{Ding2018UsingVolatility,Guo2014AdaptiveQuantification}. However, it is also critical to capture uncertainty of sparse travel demand using deep learning. \citet{Rodrigues2020BeyondProblems} explored uncertainty quantification by training a CNN-LSTM model to fit both the mean and the quantiles. They showed the power of combining the spatial and temporal embedding to predict various distributions, but they did not investigate the sparse scenario. In non-DL models, \citet{jang2005count} showed that travel demand follows a zero-inflated negative binomial distribution. \citet{rojas2020managing} also noted that using a zero-inflation model is promising for modeling intermittent travel demand. However, recent demand prediction papers sidestepped this challenge by choosing a low resolution, such as 60 minutes \citep{ke2021predicting,Ke2020}. It is challenging to consider the higher resolution, such as 5min, because a deterministic DL model is no longer appropriate. Hence, it could be a viable alternative to combine the zero-inflated distribution and the deterministic DL to analyze the sparse O-D matrices.
  
It is also critical to design metrics to evaluate the quality of uncertainty quantification. \citet{Khosravi2011LowerIntervals} proposed mean prediction interval width and prediction interval coverage probability to quantify data uncertainty. \citet{sankararaman2013distribution} introduced a likelihood-based metric, which is also an effective indicator to measure the alignment of the predicted and ground truth data distributions. Based on the existing research gap, our paper seek to design a model that formulates the sparse travel demand prediction problem with proper spatial-temporal pattern recognition and tight prediction interval bounds.

\section{Methodology}
\label{sec:method}
\subsection{Problem Description}
Our model predicts the future expected travel demand and confidence interval of each O-D pair with $k$ time windows ahead, using $m$ origins and $u$ destinations along with the travel demand in time periods (windows) of length $T$ minutes. It is a sequence-to-sequence prediction task. Different from the previous work that treated the locations of origins or destinations as vertices, we build the O-D graph $\mathcal{G} = (V,E,A)$ where $V$ represents the O-D pair set, $E$ denotes the edge set, and $A\in \mathbb{R}^{|V|\times|V|}$ is the adjacency matrix describing the relationship between O-D pairs \cite{ke2021predicting}. It is clear that $|V|=m\times u$, and the O-D graph is fully connected. Let $x_{it}$ denote the trips of the $i^{th}$ O-D pair in the $t^{th}$ time window, where $i\in V$, $x_{it}\in \mathbb{N}$. Then $X_{t}\in \mathbb{N}^{|V|\times T}$ denote the demand for all O-D pairs in the $t^{th}$ time window, with $x_{it}$ as its entry. Our goal is to leverage historical records $X_{1:t}$ as the data inputs to predict the distribution of $X_{t:t+k}$ (i.e. the demand for the next $k$ time windows), thus analyzing the expectation and confidence intervals of the future demand. 

\begin{figure*}[t]
    \centering
    \includegraphics[width=0.7\linewidth]{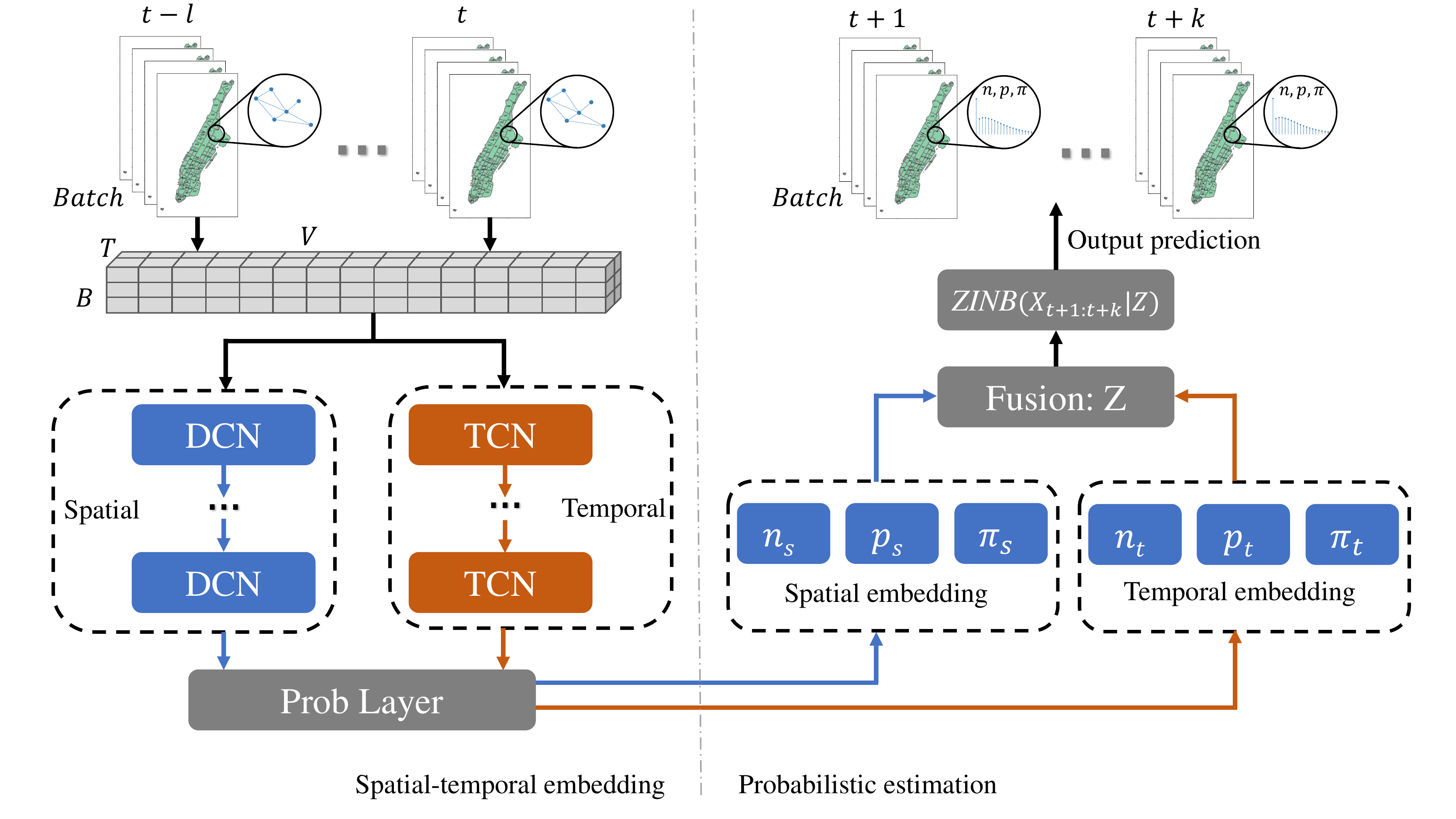}
    \caption{Framework of STZINB-GNN model.}
    \label{fig:model}
\end{figure*}

\subsection{Zero-Inflated Negative Binomial (ZINB) Distribution}
We assume that the inputs follow the ZINB distribution \citep{jiang2018sparse,rojas2020managing}. A random variable that follows NB distribution has a probability mass function $f_{NB}$ as:
\begin{equation}
    f_{NB}(x_k;n,p) \equiv Pr(X=x_k) = \left( \begin{array}{c}
        x_k+n-1   \\
        n-1
    \end{array} \right) (1-p)^{x_k}p^n.
\end{equation}
\noindent where $n$ and $p$ are the shape parameters that determine the number of successes and the probability of a single failure respectively. However, the real-world data often have many observations with zeros and overdispersion \citep{lambert1992zero}. The exploded number of zeros exacerbates the parameter learning of the NB distribution. A new parameter $\pi$ is therefore introduced to learn the inflation of zeros, leading to the ZINB distribution. Formally, its probability mass function can be described as:
\begin{equation}
    \label{eq:zinb}
    f_{ZINB}(x_k;\pi,n,p) = \left\{ \begin{array}{cc}
        \pi + (1-\pi)f_{NB}(0;n,p) & \text{if } x_k=0  \\
        (1-\pi)f_{NB}(x_k;n,p) & \text{if } x_k>0
    \end{array}\right..
\end{equation}
Two steps are needed to generate the distributions of the data: they are either zeros with probability $\pi$ or non zeros with probability $1-\pi$ following the NB distribution. The parameters $\pi,n,p$ in the probability distributions are parameterized by spatial-temporal GNNs. 


\subsection{STZINB-GNN}

We introduce STZINB-GNN, a generalizable deep learning architecture, to capture spatial-temporal correlations $Z = \mathcal{M}(X_{1:t},A)$ under the ZINB assumption for each O-D matrix entry and predict the future $k$ time windows ahead with $f_{ZINB}(X_{t+1:t+k}|Z)$. $\mathcal{M}$ is the proposed model framework that takes the historical demand $X_{1:t}\in \mathbb{Z}^{|V| \times tT}$ and adjacency matrix $A \in \mathbb{R}^{|V|\times |V|}$ as inputs, and learns the parameter embedding $Z\in \mathbb{R}^{|V|\times k \times 3}$ of the future demand $X_{t+1:t+k}\in \mathbb{R}^{|V|\times k}$ with: 
\begin{equation}
    \begin{aligned}
        &f_{ZINB}(X_{t+1:t+k}|n_{t+1:t+k},p_{t+1:t+k},\pi_{t+1:t+k}) \\
        &= f_{ZINB}(X_{t+1:t+k}|\mathcal{M}(X_{1:t},A)) = f_{ZINB}(X_{t+1:t+k}|Z).
    \end{aligned}
\end{equation}


\noindent The overall architecture of the STZINB-GNN is shown in Figure \ref{fig:model}. We convert a batch of size $B$ of input O-D-T tensor at $j^{th}$ time window $X_{j:j+B}\in \mathbb{Z}^{|V| \times BT}$ into a tensor $\mathcal{X}_{j} \in \mathbb{Z}^{|V|\times B \times T}$ as the model input. We use Diffusion Graph Convolution Networks (DGCNs) to capture the spatial adjacency of O-D pairs, and use Temporal Convolutional Networks (TCNs) for temporal correlation. The outputs of the DGCNs and TCNs, including the spatial embedding $n_s,p_s,\pi_s$ and temporal embedding $n_t,p_t,\pi_t$, are used to parameterize the ZINB distribution. These spatial and temporal embeddings contain the independent estimation of the ZINB parameters of their spatial and temporal locality. We then fuse the $n_s,p_s,\pi_s$ and $n_t,p_t,\pi_t$ into $Z$ using the Hadamard product, which can be replaced by other non-linear operations like a fully-connected layer. By fusing the spatial and temporal embeddings of the distribution parameters, we obtain the final ZINB parameter set $Z$ that fulfills
\begin{equation}
    f_{ZINB}(X_{t+1:t+k}|n_{t+1:t+k},p_{t+1:t+k},\pi_{t+1:t+k}) = f_{ZINB}(X_{t+1:t+k}|Z).
\end{equation}
\noindent Notice that the fused $Z$ in Figure \ref{fig:model} can be interpreted as the parameter set of the future demand distribution. This procedure is similar to the variational autoencoders (VAEs) concept, where we learn the latent shape variables that formulate the distributions \citep{kipf2016variational,hamilton2020graph,kingma2013auto}. The theory of VAEs is used to introduce deep learning techniques into the statistical domain of variational inference, which provides more powerful representation of latent variables. Our encoding component uses the spatial-temporal embedding architecture with an additional sparsity parameter and the decoding part is related to the probabilistic estimation of future demand.


To address the problem that zeros will give infinite values for the KL-divergence based variational lower bound, we directly use the negative likelihood as our loss function to better fit the distribution into the data. Let $y$ be the ground-truth values corresponding to one of the predicted matrix entries with parameters $n,p,\pi$ from $Z$. The log likelihood of ZINB is composed of the  $y=0$ and $y>0$ parts, and can be approximated as \citep{minami2007modeling,moghimbeigi2008multilevel}:
\begin{equation}
    LL_y=\left\{ 
        \begin{array}{lc}
            \log{\pi} + \log{(1-\pi)p^n} & \text{when}\quad y=0 \\
            \log{1-\pi} + \log{\Gamma(n+y)} - \log{\Gamma(y+1)} & \\
            \quad\quad -\log{\Gamma(n)} + n\log{p} + y\log(1-\pi) & \text{when}\quad y>0
        \end{array}
    \right.,
\end{equation}
\noindent where $\pi,n,p$ are also selected and calculated according to the index of $y=0$ or $y>0$, and $\Gamma$ is the Gamma function. The final negative log likelihood loss function is given by:
\begin{equation}
    NLL_{STZINB} = - LL_{y=0} - LL_{y>0}.
\end{equation}
Note that our model can also be generalized to other distributions by modifying the probability layer into other distributions and using $Z$ to represent the related shape parameter sets. For example, if we use Gaussian distribution as our model, we can parameterize the probability layer using the spatial and temporal embedding of the mean and variance, thus quantifying the data uncertainty that follows the Gaussian distribution. Using the flexibility of the probability layer, we design a variety of benchmark models to compare to the ZINB model. Our scripts can be found in Github\footnote{\url{https://github.com/ZhuangDingyi/STZINB}}.

\subsection{Adjacency Matrix for O-D Pairs}
\label{sec:adj}
This study uses O-D pairs as vertices, different from the previous GNN approaches that use regions as vertices \citep{Geng2019SpatiotemporalForecasting,sun2020predicting,geng2019multi,yao2018deep}. Therefore, we need to model spatial correlations of O-D pairs and construct the adjacency matrix in a different way. Intuitively, the O-D pairs with similar origins or destinations are also close in a graph representation, because passengers are likely to transfer between adjoining O-D pairs rather than remote ones. Inspired by the work of \citet{ke2021predicting}, we formulate our adjacency matrix as:
\begin{equation}
    \begin{aligned}
    A^O_{i,j} &= haversine(lng^O_i,lat^O_i,lng^O_j,lat^O_j)^{-1},\forall i,j \in V\\
    A^D_{i,j} &= haversine(lng^D_i,lat^D_i,lng^D_j,lat^D_j)^{-1},\forall i,j \in V\\
    A_{i,j} &= \sqrt{\frac{1}{2} ((A^O_{i,j})^2 + (A^D_{i,j})^2) },
    \end{aligned}
\end{equation}

\noindent where $lng^O_i,lat^O_i,lng^O_j,lat^O_j$ are the longitudes and latitudes of the origins of O-D pair $i,j$, and $lng^D_i,lat^D_i,lng^D_j,lat^D_j$ are for the destinations similarly. The function $haversine(\cdot)$ takes the longitudes and latitudes of two geographical points and calculate their distance on Earth. The basic idea is to leverage the O-D pairs' geographical adjacency by averaging the origin and destination similarity. It is clear that the order of $i,j$ does not affect the output of the $haversine$ function, which means $A$ is symmetric. The final adjacency matrix $A$ is the quadratic mean of $A^O$ and $A^D$, where the distances between origins or destinations have the same influence in the adjacency matrix. Future studies can assign different weights to the origins or destinations in constructing the adjacency matrix, or even combine with demographic graphs to enrich the information of $A$. Since this paper focuses on uncertainty quantification and interpretability of the model, we use a simple construction of $A$ to prevent any distraction.

\subsection{Diffusion Graph Convolution Network} 

In order to capture the stochastic nature of flow dynamics among O-D pairs, we model the spatial correlations as a diffusion process \citep{li2017diffusion,atwood2016diffusion}. Introducing the diffusion process facilitates the learning of the spatial dependency from one O-D pair to another. The process is characterized by a random walk on the given graph with a probability $\alpha \in [0,1]$ and a forward transition matrix $\Tilde{W}_f = A/rowsum(A)$ \citep{wu2021inductive}. After sufficiently large time steps, the Markovian property of the diffusion process guarantees it to converge to a stationary distribution $\mathcal{P}\in \mathbb{R}^{|V|\times |V|}$. Each row of $\mathcal{P}$ stands for the probability of diffusion from that node. The stationary distribution can be calculated in closed form \citep{teng2016scalable}:
\begin{equation}
    \mathcal{P} = \sum_{k=0}^{\infty} \alpha (1-\alpha)^k (D_O^{-1}A)^k,
\end{equation}
\noindent where $k$ is the diffusion step, which is usually set to finite number $K$. Variable $k$ is only reused for demonstration of the diffusion process. We can similarly define the backward diffusion process with backward transition matrix $\Tilde{W}_b = A^T/rowsum(A^T)$. Since our adjacency matrix $A$ is symmetric, $\Tilde{W}_f = \Tilde{W}_b$. The forward and backward diffusion processes model the dynamics of passengers shifting from one O-D pair to another, like the shifting from school-home trip to home-market trip. The building block of DGCN layer can be written as \citep{wu2021inductive}:
\begin{equation}
    H_{l+1} = \sigma(\sum_{k=1}^{K} T_k (\Tilde{W}_f)H_l \Theta^{k}_{f,l} + T_k (\Tilde{W}_b)H_l \Theta^k_{b,l}),
\end{equation}
\noindent where $H_l$ represents the $l^{th}$ hidden layer; the Chebyshev polynomial $T_k(X) = 2XT_{k-1}(X) - T_{k-2}(X)$ is used to approximate the convolution operation in DGCN, with boundary conditions $T_0(X) = I$ and $T_1(X) = X$ (note that the Chebyshev polynomial is used to approximate the diffusion $\mathcal{P}$ instead of using the closed form); learned parameters of the $l^{th}$ layer $\Theta^{k}_{f,l}$  and $\Theta^k_{b,l}$ are added to control how each node transforms the received information; $\sigma$ is the activation function (e.g. ReLU, Linear). In our model we stack 3 DGCN layers to better capture the O-D dynamics.

\subsection{Temporal Convolutional Network}

As \citet{wu2021spatial} point out, the advantages of TCNs compared with recurrent neural networks (RNNs) include: 1) TCNs can use the sequences with varying length as inputs, which is more adaptive to different time resolutions and scales; 2) TCNs have a lightweight architecture and fast training \citep{gehring2017convolutional}. The general idea of TCNs is to apply a shared gated 1D convolution with width $w_l$ in the $l^{th}$ layer in order to pass the information from $w_l$ neighbors at the current time point. Each TCN layer $H_{l}$ receives the signals from the previous layer $H_{l-1}$ and is updated using \citep{lea2017temporal}:
\begin{equation}
    H_{l} = f(\Gamma_l*H_{l-1} + b),
\end{equation}
\noindent where $\Gamma_l$ is the convolution filter for the corresponding layer, $*$ is the shared convolution operation, and $b$ stands for the bias. If previous hidden layer follows $H_{l-1}\in \mathbb{R}^{B\times |V| \times w_{l-1}}$, then the convolution filter is $\Gamma_l\in \mathbb{R}^{w_{l}\times w_{l-1}}$ so that $H_{l} \in \mathbb{R}^{B\times |V| \times w_{l}}$. If there is no padding in each TCN layer, it is clear that $w_{l}<w_{l-1}$. TCN is also a type of sequence-to-sequence model, which can directly output prediction for future target windows in a row. Furthermore, the TCN receptive field is flexible, which can be controlled by the number and kernels of $\Gamma_l$. It is useful in our following discussion about scaling our model with different temporal resolutions. We also stack 3 TCN layers.

\section{Numerical Experiments}
\label{sec:experiment}
\subsection{Data}
In this session, we assess the model performance using two real-world datasets from Chicago Data Portal (CDP) \footnote{\url{https://data.cityofchicago.org/Transportation/Transportation-Network-Providers-Trips/m6dm-c72p}} and Smart Location Database (SLD) \footnote{\url{https://www1.nyc.gov/site/tlc/about/tlc-trip-record-data.page}}. The \textbf{CDP dataset} contains the trip records of Transportation Network Providers (ride-sharing companies) in the Chicago area. The city of Chicago is divided into 77 zones and the trip requests with pick-up and drop-off zone are recorded every 15min. We use 4-month observations from September 1st, 2019 to December 30th, 2019. The dataset is divided into various spatial resolutions to facilitate the discussion of model performance. We randomly select $10\times 10$ O-D pairs with the same time period as spatially sparse data sample; The \textbf{SLD dataset} is also used in the work of \citet{ke2021predicting}. Specifically, we select the For-Hire Vehicle (FHV) trip records in the Manhattan area (divided into 67 zones by administrative zip codes) from January 2018 to April 2018. This SLD data have similar features as the CDP dataset, including timestamps, and pick-up and drop-off zone IDs. As the dataset includes the timestamps of the individual trip information, we vary the temporal resolution (5min, 15min, and 60min intervals) to test our model performance. In addition to the full O-D matrix ($67\times 67$), we also sample $10\times 10$ small O-D pair samples to align with the CDP dataset. We use 60\% of the data for training, 10\% for validation, and the last 30\% for testing. Table \ref{tab:data_description} summarizes the data scenarios used in the analysis.

\begin{table}[htbp]
    \centering
    \small
    \begin{tabular}{c|cccc}
    \hline
        Name & Resolution & \# of O-D pairs & Data size & Zero rate \\
    \midrule
        CDP\_SAMP10 & 15min & $10\times 10$ & $(100,11521)$ & 81\%  \\
        SLD\_SAMP10 & 15min & $10\times 10$ & $(100,11520)$ & 54\%  \\
        SLD\_5min   & 5min  & $67\times 67$ & $(4489,34560)$ & 88\% \\
        SLD\_15min  & 15min & $67\times 67$ & $(4489,11520)$ & 70\% \\
        SLD\_60min  & 60min & $67\times 67$ & $(4489,2880)$  & 50\% \\
    \hline
    \end{tabular}
    \caption{Data division summary.}
    \label{tab:data_description}
\end{table}

\begin{figure}[!t]
    \centering
    \includegraphics{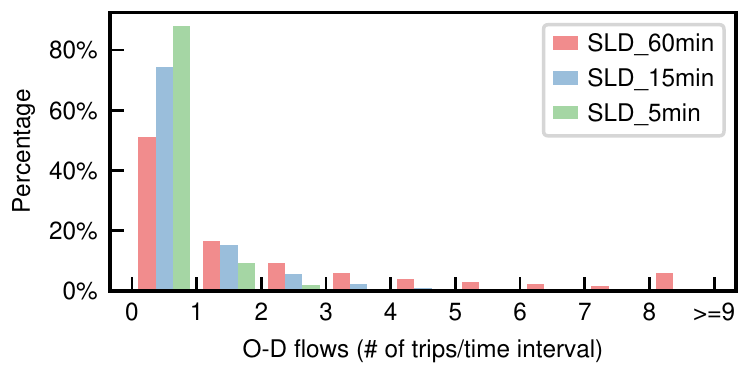}
    \caption{Distribution of travel demand in the SLD dataset per 5, 15, and 60min intervals.}
    \label{fig:sld_data}
\end{figure}

The zero rate increases from 50\% to 88\% as the temporal resolution for SLD dataset increases from 60 minutes to 5 minutes. As shown in Figure \ref{fig:sld_data}, the distribution of O-D flows is quite skewed because most of the O-D flows are concentrated around small values. The O-D flows at a 60min resolution have several observations larger than 8, but at 5min resolution, the lowest value is 3 trips within the 5-minute interval. Such sparse (and small value) O-D matrix is a challenge for most of the deterministic deep learning models.


\subsection{Evaluation Metrics}
We use the metrics for both the point estimate statistics and distributional characteristics to evaluate and compare the performance of the various models. The prediction accuracy of the expected median value (i.e. point estimate accuracy) is evaluated using the Mean Absolute Error (MAE): 

\begin{equation}
\text{MAE}=\frac{1}{k|V|} \sum_{i=1}^{k|V|} \left|x_i - \hat{x}_{i}\right|,
\end{equation} 

\begin{table*}[!h]
\footnotesize
\caption{Model comparison under different metrics. $X/Y$ values correspond to the mean/median values of the distribution.}
\label{tab:comparison}
\centering
\begin{tabular}{cc|c c c c c c}
\toprule
Data scenario&Metrics & STZINB-GNN & STNB-GNN & STG-GNN & STTN-GNN & HA & STGCN\\
\midrule
\multirow{5}{*}{CDP\_SAMP10}&MAE           & 0.368/\textbf{0.366} & 0.382/0.379 & 0.409/0.409 & 0.432/0.606 & 0.522 & 0.395\\
&MPIW      & \textbf{1.018} & 1.020 & 2.407 & 2.089 & /\ & /\\\
&KL-Divergence & \textbf{0.291}/0.424 & 0.342/0.478 & 0.435/0.435 & 1.058/0.928 & 1.377\ & 0.897\\
&True-zero rate & 0.796/0.788 & 0.796/0.788 & 0.790/0.790 & 0.758/0.764 & 0.759 & \textbf{0.800}\\
&F1-Score  & \textbf{0.848}/0.846 & \textbf{0.848}/0.841 & 0.818/0.818 & 0.842/0.846 &  0.809 & 0.840\\

\midrule

\multirow{5}{*}{SLD\_SAMP10}&MAE & 0.663/0.666 & 0.627/\textbf{0.616} & 0.630/0.630 & 0.695/0.665  & 0.697 & 0.630\\
&MPIW      & \textbf{1.310} & 3.628 & 2.604 & 1.931  & /\ & /\ \\
&KL-Divergence & 0.518/\textbf{0.507} & 0.980/1.662 & 1.022/1.022 & 3.578/3.052  & 0.978 & 0.768\\
&True-zero rate & 0.499/\textbf{0.502} & 0.465/0.418 & 0.461/0.461 & 0.308/0.336  & 0.364 & 0.478\\
&F1-Score  & \textbf{0.567}/0.566 & 0.556/0.552 & 0.555/0.555 & 0.477/0.500  & 0.456 & 0.563\\

\midrule

\multirow{5}{*}{SLD\_5min}&MAE & 0.149/0.150 & 0.147/\textbf{0.144} & 0.155/0.155 & 0.155/0.155 & 0.149 & 0.159\\
&MPIW      & \textbf{0.094} & 1.249 &  0.922 & 0.741  & /\ &  /\ \\
&KL-Divergence  & 0.015/0.014  & 0.042/0.145 & \textbf{0.001/0.001} & 0.001/0.001  &  0.060 & 0.056 \\
&True-zero rate & \textbf{0.879/0.879} & 0.875/0.866 & 0.877/0.877 & 0.877/0.877 &  0.874 & 0.874 \\
&F1-Score & \textbf{0.882/0.882} & 0.880/0.878 & 0.879/0.879 & 0.879/0.879 & 0.876  & 0.879 \\

\midrule

\multirow{5}{*}{SLD\_15min}&MAE & 0.370/0.372 & 0.351/\textbf{0.342} & 0.356/0.356 & 0.365/0.356 &     0.418 & 0.373\\
&MPIW       & \textbf{0.603} & 2.283 & 1.353 &  1.215   & /\ & /\ \\
&KL-Divergence  & 0.167/\textbf{0.156} & 0.357/0.704 & 0.353/0.353 &  1.445/1.211 & 0.445 & 0.395\\
&True-zero rate & 0.725/\textbf{0.727} & 0.710/0.684 & 0.709/0.709 &  0.632/0.648 & 0.703 & 0.708\\
&F1-Score       & \textbf{0.751}/0.750 & 0.746/0.745 & 0.750/0.750 &  0.716/0.726 & 0.744 & 0.750\\

\midrule

\multirow{5}{*}{SLD\_60min}&MAE & 1.040/1.067 & 0.958/\textbf{0.947} & 1.199/1.199 & 1.275/1.254  & 1.014 & 0.997\\
&MPIW      & 3.277 & 5.753 & 2.282 & \textbf{1.592}  & /\ & /\ \\
&KL-Divergence  & 0.982/1.270 & \textbf{0.926}/0.963 & 2.176/2.176 & 4.120/3.734 &  2.421  & 1.114 \\
&True-zero rate & 0.458/\textbf{0.476} & 0.443/0.425 & 0.390/0.390 & 0.288/0.308 &  0.447  & 0.438 \\
&F1-Score       & 0.536/0.537 & \textbf{0.538}/0.534 & 0.479/0.479 & 0.407/0.423 &   0.490 & \textbf{0.538} \\

\bottomrule
\end{tabular}
\end{table*}

\begin{figure*}[!h]
    \centering
    \subfigure[SLD\_15min]{\includegraphics{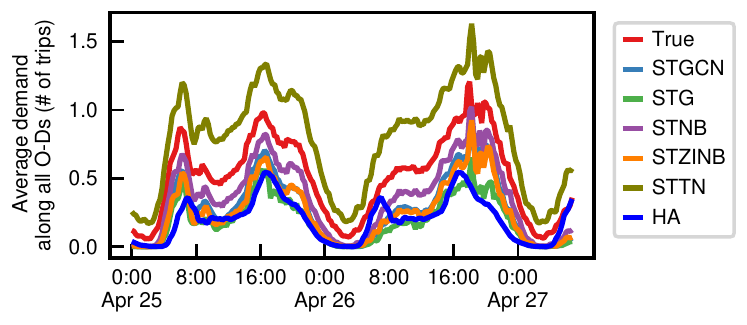}}
    \subfigure[SLD\_60min]{\includegraphics{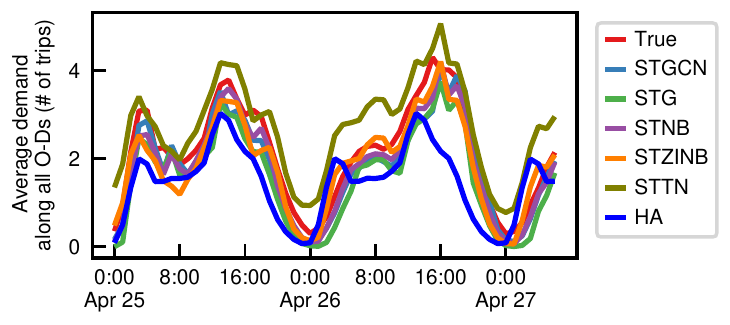}}
    \caption{Prediction results in the SLD\_15min and SLD\_60min scenarios. Results are averaged over the spatial dimension (i.e. all O-D pairs)}
    \label{fig:performance}
\end{figure*}

\begin{figure*}[b]
    \centering
    \subfigure[SLD\_5min case]{\includegraphics{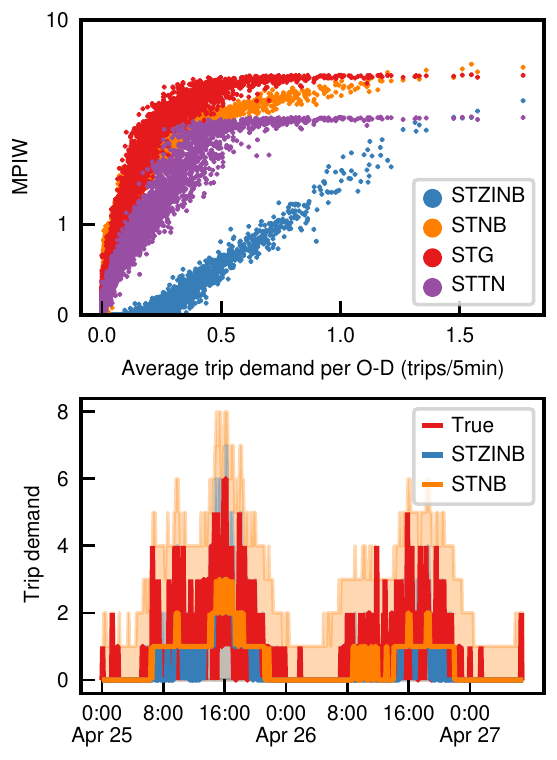}}
    \subfigure[SLD\_15min case]{\includegraphics{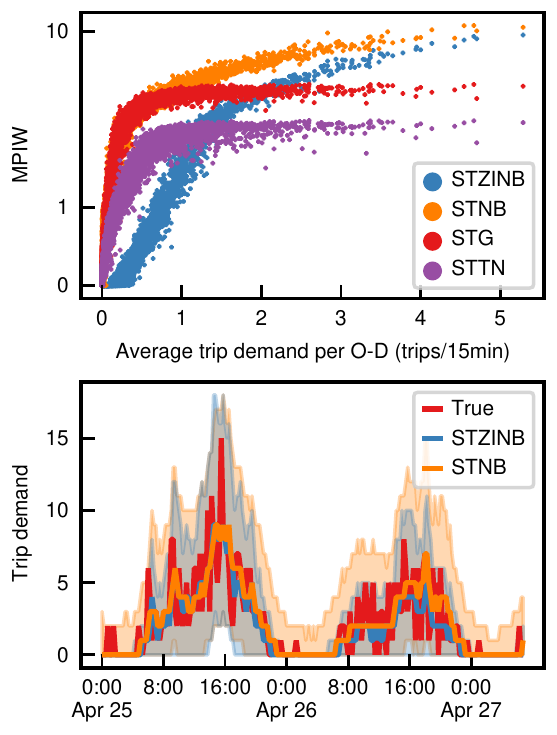}}
    \subfigure[SLD\_60min case]{\includegraphics{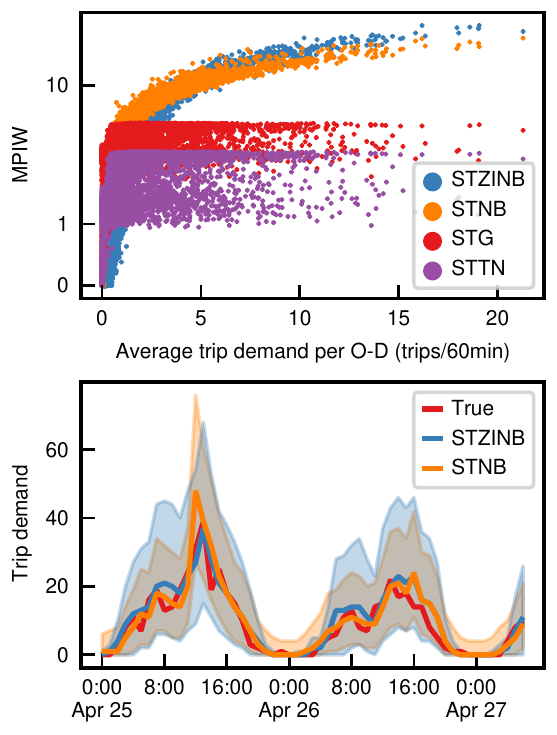}}
    \caption{Model prediction uncertainty at various resolutions (SLD\_5min to SLD\_60min). The top row shows the MPIW needed to predict the demand of each O-D pair at different resolutions. The bottom row represents the prediction with uncertainty for a specific O-D pair, from Midtown Center to Union Square, in April 25 and 26, 2018. This O-D pair is selected because it has the largest trip flow.}
    \label{fig:uncertainty}
\end{figure*}

\noindent where $\hat{x}_i$ and $x_{i}$ are the predicted and ground-truth values of $i^{th}$ data point respectively. To evaluate the estimated uncertainty, we use Mean Prediction Interval Width (MPIW) on the 10\%-90\% confidence interval \citep{Khosravi2011LowerIntervals}:
\begin{equation}
\text{MPIW}=\frac{1}{k|V|} \sum_{i=1}^{k|V|} (U_i - L_i).
\end{equation} 
where $L_i$ and $U_i$ correspond to the lower and upper bound of the confidence interval for observation $i$. The definition of MPIW can be extended to other quantiles of the data. Tighter (smaller) prediction intervals are more desirable. Apart from MPIW, we also use the Kullback-Leibler Divergence (KL-Divergence) to assess how close the model output distribution is to the test set distribution \citep{kullback1951information}. We define the KL-Divergence as:
\begin{equation}
\text{KL-Divergence}=\frac{1}{k|V|} \sum_{i=1}^{k|V|} (\hat{x}_i \log\frac{\hat{x}_i+\epsilon}{x_i+\epsilon} ),
\end{equation} 
Since many $x_i$ and $\hat{x}_i$ values are likely to be zeros, a small perturbation $\epsilon = 10^{-5}$ is used to avoid numerical issue because of division by 0. Since the KL-Divergence measures the difference between two distributions, smaller values are desirable. 

To compare the model performance on the discrete O-D matrix entries, we use the true-zero rate and F1-score measurements. The true-zero rate quantifies how well the model replicates the sparsity in the ground-truth data. The F1-score, on the other hand, measures the accuracy of discrete predictions. Even though the F1-score is designed for classification models, we can still consider the discrete values as multiple labels and define the precision and recall accordingly \citep{pedregosa2011scikit}. Larger true-zero rate and F1-score values indicate better model performance.


\subsection{Model Comparison}
\label{sec:model_comparison}

In order to explore the advantages of the STZINB-GNN, we compare the STZINB-GNN results against three other models: (1) \textbf{Historical Average (HA)} serves as the statistic baseline. It is calculated by averaging the demand in the same daily time intervals (e.g. 8:00AM-8:15AM) from the historical data to predict the one-step ahead future demand. (2) \textbf{Spatial-Temporal Graph Convolutional Networks (STGCN)}\footnote{\url{https://github.com/FelixOpolka/STGCN-PyTorch}} is the state-of-the-art deep learning model for traffic prediction \cite{Yu2018Spatio-TemporalForecasting}. STGCN also uses graph convolution for spatial embedding and temporal convolution for temporal embedding. However, it fails to quantify the demand uncertainty and only produces point estimates; (3) \textbf{Models with probabilistic assumptions} in the spatial-temporal embedded probabilistic layer: negative binomial (STNB-GNN), Gaussian (STG-GNN), and truncated normal (STTN-GNN), as shown in Figure \ref{fig:model}. These three distributions have two parameters, less than the three parameters of STZINB-GNN. The other components and parameters are the same across these models. 

STZINB-GNN outperforms other models when the O-D matrix resolution is high but performs worse when the resolution becomes coarser. Table~\ref{tab:comparison} highlights the fields in bold to indicate the best performance for each data set. For fair comparison, the outputs from the continuous-output models, including STG-GNN, STTN-GNN, HA, and STGCN, were rounded to the closest integer to compare to the STNB-GNN and STZINB-GNN. As shown in Table~\ref{tab:comparison}, when the spatial and temporal resolutions are low, like the CDP\_SAMP10 and SLD\_SAMP10 cases, STZINB-GNN performs similarly to STNB-GNN. In the sparsest scenario, SLD\_5min (with 88\% zeros), STG-GNN, STTN-GNN, and STGCN fail to effectively capture the skewed data distribution, leading to low true-zero rates and F1-scores. The STZINB-GNN, on the other hand, successfully learns the sparsity of the data. Its prediction has a very small MPIW, nearly 10 times smaller than the other models. When the temporal resolution decreases, the STNB-GNN, STTN-GNN, and STGCN start to outperform STZINB-GNN. In the SLD\_60min case, STNB-GNN fits the data better than the STGCN. The STZINB-GNN, on the other hand, only predicts the true-zero entries better. No single model dominates others under all resolution levels \citep{wolpert1997no}.


The negative binomial distribution in the probability layer can effectively capture the discrete values of the travel demand. Figure \ref{fig:performance} illustrates the predicted and ground-truth average travel demand in the SLD (New York) data for two consecutive days, with Figure \ref{fig:performance}(a) for the SLD\_15min and Figure \ref{fig:performance}(b) for the SLD\_60min. With finer resolutions, the STNB-GNN and STZINB-GNN models are more likely to accurately predict the average travel demand. When the resolution decreases, like in the 60min case, all the deep learning models deliver similar performance. This 60-min resolution is commonly used in the majority of the deep learning studies, but our results demonstrate the importance of discrete probabilistic assumptions when the temporal unit is shorter than 60 minutes.

\subsection{Uncertainty Quantification}
\label{sec:uq}
The STZINB-GNN provides superior performance in quantifying uncertainty, because it uses much narrower MPIW for a fixed probability coverage, particularly when the temporal resolution is high. Figure \ref{fig:uncertainty} visualizes the scatter plots for the MPIW and the groundtruth travel demand of the 4489 O-D pairs at three temporal resolutions. Figure \ref{fig:uncertainty}(a) demonstrates that the STZINB-GNN leads to significantly smaller MPIW than the other models in the granular 5min resolution case. This is because the average travel demand for each O-D pair is less than two in the SLD\_5min case and most of them are zeros. The introduction of the sparsity parameter $\pi$ in STZINB-GNN effectively captures the zeros. The STG-GNN and STTN-GNN are not able to capture the skewness of the data distribution, leading to large MPIWs. However, when the temporal resolution decreases, the zero-inflation with the sparsity parameter $\pi$ becomes less important. Figures \ref{fig:uncertainty}(b) and \ref{fig:uncertainty}(c) show that the SLD\_15min and SLD\_60min cases have larger average travel demand per 15 and 60 minutes time interval. In these two cases, the STZINB-GNN has smaller MPIW only when the average travel demand is small. The STG-GNN and STTN-GNN outperform STZINB-GNN when the average demand is large. However, their output prediction distributions are not stable, because the same travel demand value corresponds to different MPIW with a large variance.

The bottom row of Figure \ref{fig:uncertainty} compares the expectation and MPIW between STZINB-GNN and STNB-GNN for a specific O-D pair that has the largest demand flow in the selected time range. In the SLD\_5min and SLD\_15min cases, the STZINB-GNN provides more compact confidence intervals than the STNB-GNN, even when its point prediction is less accurate. The results are consistent with the results in Table \ref{tab:comparison} and the MPIW comparison in the top row of Figure \ref{fig:uncertainty}. 


\subsection{Interpretation of the Sparsity Parameter $\pi$}
\begin{figure}[htbp]
    \centering
    \subfigure[Morning peak]{\includegraphics{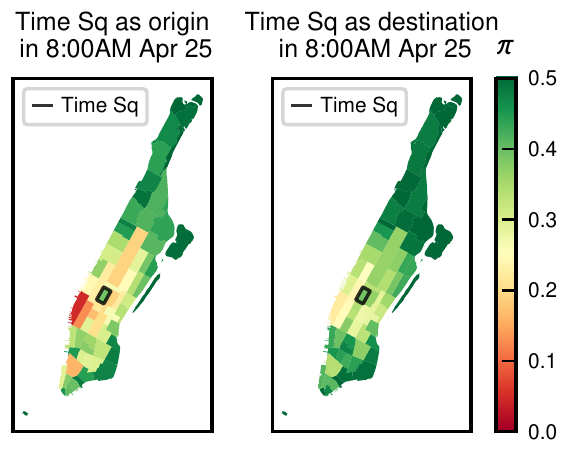}}
    \subfigure[Evening peak]{\includegraphics{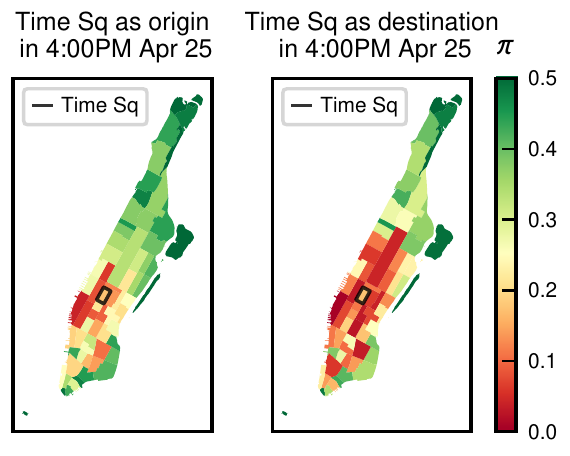}}
    \caption{Morning and evening peaks outflow from and inflow to Time Square. We project the O-D pairs into the map according to Time Square as the origin or the destination zone. Red zones stand for small $\pi$ values, meaning high possibility to have trips generated there and green regions otherwise. }
    \label{fig:region}
\end{figure}

The sparsity parameter $\pi$ in the ZINB distribution measures how likely a zone has zero demand. Note that each of our predicted O-D pair has the parameter $\pi$, which can capture the inflow and outflow sparsity level of a zone. Since O-D relationship is hard to illustrate in the map, we focus on a specific zone and project the O-D activities when the zone is selected as the origin or destination. Figure \ref{fig:region} shows the O-D activity (i.e. the parameter $\pi$) heatmap of Time Square, during the morning and evening peaks of April 25, 2018. It can be found that spatial locality exists, where communities are more likely to commute to their neighbors. Moreover, the temporal patterns also vary. As shown in Figure \ref{fig:region}, many visitors explore the neighboring regions in the evening but are very inactive during the morning peak. Therefore, the sparsity parameter $\pi$ renders the STZINB-GNN highly interpretable. It is very important in the transportation decision-making or the operation manager to use the sparsity parameter $\pi$ to assign mobility service. Our framework has the potential extended to other prediction tasks that have many zeros and are sensitive to events, like incident precaution or paratransit service for the disabled people.

\section{Conclusion}
\label{sec:conclusion}

In this paper, we propose a generalizable spatial-temporal GNN framework to predict the probabilistic distribution of sparse travel demand and quantify its uncertainty. We introduce the zero-inflated negative binomial distribution with a sparsity parameter $\pi$. We use spatial diffusion graph neural networks to capture spatial correlation and temporal convolutional networks to capture temporal dependency. The STZINB-GNN framework embeds the spatial and temporal representation of the distribution parameters respectively and fuses them to obtain the distribution for each spatial-temporal data point. The STZINB-GNN is evaluated using two real-world datasets with different spatial and temporal resolutions. We find that the STZINB-GNN outperforms the baseline models when the data are represented in high resolutions but performs worse when the resolution becomes coarser. This is also reflected in the prediction interval, where the STZINB-GNN has tight confidence intervals. Since the parameter $\pi$ has physical interpretation, our model could help transportation decision makers to efficiently assign mobility services to zero or non-zero demand areas. This framework has the potential to be extended to other prediction tasks that use highly sparse data points, such as anomaly detection and accident prediction.

\begin{acks}
This material is based upon work supported by the U.S. Department of Energy’s Office of Energy Efficiency and Renewable Energy (EERE) under the Vehicle Technology Program Award Number DE-EE0009211.
\end{acks}


    


\bibliographystyle{ACM-Reference-Format}
\bibliography{acmart}

\end{document}



\renewcommand{\shortauthors}{XXX and XXX, et al.}






\maketitle

\section{Data Description}

We use seven real-world spatiotemporal datasets to evaluate our model (see Table \ref{tab:data} for an overview). They are:

\textbf{METR-LA}\footnote{\url{https://github.com/liyaguang/DCRNN}} consists of traffic speed information collected by highway loop detectors in Los Angeles \citep{li2018diffusion,wu2020inductive}. We follow the experiment settings of \citet{li2018diffusion} to select 4 months data from Mar 1st 2012 to Aug 30th 2012 with 207 sensors. 

\textbf{NREL}\footnote{\url{https://www.nrel.gov/grid/solar-power-data.html}} contains many energy datasets, and here we choose the Alabama Solar Power Data for Integration Studies \citep{sengupta2018national}. This dataset includes 5-minute solar power records of 137 photovoltaic power plants in 2006. We follow the work of \citet{wu2020inductive} by only keeping the data from 7 am-7 pm everyday in order to attenuate this effect.

\textbf{PeMS-Bay}\footnote{\url{https://github.com/liyaguang/DCRNN}} is also a traffic speed dataset that is collected in Bay Area by Performance Measurement System (PeMS). Same as the work of \citet{li2018diffusion}, we choose 325 sensors from Jan 1st 2017 to May 13th 2017.

\textbf{NOAA}\footnote{\url{https://github.com/MengyangGu/GPPCA}} records the global gridded air and marine temperature anomalies from U.S. National Oceanic and Atmospheric Administration (NOAA) \citep{gu2020generalized}. NOAA contains the monthly data from Jan 1999 to Dec 2018 with $5^{\circ}\times 5^{\circ}$ latitude-longitude resolution. We follow the work of \citet{gu2020generalized} to leave out polar circles and choose 1639 observed grids. With the largest 20 singular values take only 46.6\% portion, the dataset is hard to be considered low-rank.

\textbf{MODIS}\footnote{\url{https://modis.gsfc.nasa.gov/data/}} consists of daytime land surface measured by the Terra platform on the MODIS satellite with 3255 downsampled grids from Jan 1, 2019 to Jan 16, 2021. It is automatically collected by \textit{MODIStsp} package in \textit{R} \citep{busetto2016modistsp}. We follow the work of \citet{heaton2019case} to select a region with similar longitude-latitude bounds as the target (latitude: 30.5$\sim$37.6; longitude: -96.8$\sim$-89.7). Because of the cloud covering the satellites, there are more than 39.6\% entries are missing in the collected data. Moreover, the missing data are generally the continuous spatial area, which is really challenging for models to train.

\textbf{USHCN}\footnote{\url{https://www.ncdc.noaa.gov/ushcn/introduction}} contains monthly precipitation of 1218 locations from 1899 to 2019, which is collected by the U.S. Historical Climatology Network (USHCN) \citep{menne2009us}. As \citet{wu2020inductive} mentioned, USHCN dataset is pretty dispersed, with variance-to-mean ratio exceeds 500. Thus it can help examine the model performance on time series with substantial oscillations.


We summarize these datasets in Table \ref{tab:data}

\begin{table}[!ht]
\small
\footnotesize
    \centering
    \caption{Real-world spatiotemporal datasets description}
    \begin{tabular}{ccccccc}
    \toprule
            & Sensors & Time length & Distance type& Normalization parameter $\sigma$  & Usage & Missing  \\
                   &  & (frequency)& &  of adjacency matrix  &  &  \\
    \midrule
    METR-LA & 207 & 34272 (5-min)& Travelling distance & 557 & SATCN training & 8.11\%\\
    NREL  & 137 & 105120 (5-min) & Haversine distance &14,000  & SATCN training & - \\
    PeMS-Bay  & 325 & 52116 (5-min)& Travelling distance & 1.5   & SATCN training & 0.003\%\\
    NOAA  & 1639 & 240 (1-month)& Haversine distance& 6,000,000   & SATCN training & -\\
    MODIS  & 3225 & 747 (1-day)& Haversine distance & 112,000   & SATCN training & 39.62\%\\
    USHCN & 1218 & 1440 (1-month)& Haversine distance & 100,000  & SATCN training & 3.07\%\\
    \bottomrule
    \end{tabular}
    \label{tab:data}
\end{table}

\section{Implementation Details of SATCN}

\section{Benchmark Settings}
\textbf{kNN} For traffic datasets METR-LA/PeMS-Bay/PeMSD4, the distance is based on the road distance in traffic network. While haversine distance is used in geospatial datasets NREL/NOAA/MODIS/USHCN. For each dataset, we tune the best $k$ according to the lowest kriging RMSE errors. The selection of $k$ for different datasets includes: \textit{METR-LA:} 7; \textit{NREL:} 10; \textit{PeMS-Bay:} 4; \textit{USHCN:} 7; \textit{NOAA:} 2; \textit{MODIS:} 3 and \textit{USHCN:} 7.

\textbf{Okriging} Ordinary kriging is a common used geostatistical method in interpolating values by fitting prior spatial covariance in Gaussian process. We use \textit{Automap} package in \textit{R} to implement the ordinary kriging \citep{krige1951statistical,carley2012automap}. To keep consistent with the construction of adjacency matrix, we fix the variogram kernels as "Gaussian" and automatically learns their parameters by \textit{autoKrige} function. Note that we apply the method at each snapshot. In other words, at each time point, we provide the corresponding values of observed nodes to the algorithm to obtain the interpolated column vector as kriging result. This task fulfills embarrassingly parallel for all time steps.

\textbf{GLTL} Greedy Low-rank Tensor Learning \citep{bahadori2014fast} is designed for both co-kriging and forecasting tasks for multiple variables. There are only one variable for our datasets. We implement this with the \textit{MATLAB} source code\footnote{\url{http://roseyu.com/code.html}} of the authors. We choose the \textit{Orthogonal} algorithm according to its superior performance in \citep{bahadori2014fast}. We set the maximum number of iterations to 1000 and the convergence stopping criteria to $1\times 10^{-10}$. For all the datasets, we use the same Gaussian kernel based adjacency matrices. The Laplacian matrix is calculated by $L=D-W$ where $D$ is a diagonal matrix with $D_{ii} = \sum_{j}W_{ij}$. Same to the implementation in \cite{bahadori2014fast}, we rescale the Laplacian matrix by ${L}=\frac{L}{\max_{ij} L_{ij}}$ as well. An essential parameter is $\mu$ for the weight of Laplacian regularizer, which is tuned by performing grid search from $\{0.05,0.5,5,50,500\}$. We reach convergence before the maximum 1000 iterations in all datasets. The tuned $\mu$ for different datasets are listed in Table \ref{tab:gltl_para}.

\begin{table}[!ht]
    \centering
    \caption{GLTL Parameters for each dataset}
    \begin{tabular}{ccccccc}
    \toprule
    Parameter & METR-LA & NREL & NOAA & MODIS & PeMS-Bay & USHCN \\
    \midrule
$\mu$ & 0.5 & 50 & 5 & 0.5 & 5 & 5  \\
    \bottomrule
    \end{tabular}
    \label{tab:gltl_para}
\end{table}

\textbf{IGNNK} Inductive Graph Neural Network for Kriging \citep{wu2020inductive} is a novel GNN based model that combines dynamic subgraph sampling techniques and diffusion graph convolution structure for the kriging task. We implement this from the Github repository\footnote{\url{https://github.com/Kaimaoge/IGNNK}} of \citep{wu2020inductive}. For datasets with distance information, it uses a Gaussian kernel to construct the adjacency matrix for GNN:
\begin{equation}
W_{ij} = \exp\left(-\left(\frac{\text{dist}\left(v_i , v_j\right)}{\sigma}\right)^2\right),
\label{distance_rule}
\end{equation}
where $W_{ij}$ stands for adjacency or closeness between sensors/nodes $v_i$ and $v_j$, $\text{dist}\left(v_i , v_j\right)$ is the distance between $v_i$ and $v_j$, and $\sigma$ is the normalization parameter, which is illustrated in Table \ref{tab:data}. We use Adam optimizer with learning rate 0.0001 to optimize the GNN training and set the maximum number of training episodes as 750. Apart from them, there are many identical model parameters defined for each dataset. We follow similar settings in the work \citet{wu2020inductive}, which are listed in Table \ref{tab:para}:

\begin{table}[!ht]
\small
    \centering
    \caption{IGNNK Parameters for each dataset}
    \begin{tabular}{ccccccc}
    \toprule
{Parameters} & METR-LA & NREL & USHCN & NOAA & PeMS-Bay & MODIS \\
    \midrule
window length $h$ & 24 & 16 & 6 & 1 & 24 & - \\
number of evaluation windows (test) & 428 & 1971 & 72 & 72 & 2171 & - \\
number of kriging nodes ($n_t^u$) & 50 & 30 & 300 & 500 & 80 & - \\
sampled observed nodes size $n_o$ & 100 & 100 & 900 & 1100 & 240 & - \\
sampled masked nodes size $n_m$ & 50 & 30 & 300 & 400 & 80 & - \\
hidden feature dimension $z$ &100 & 100 & 100 & 100 & 100 & - \\
activation function $\sigma$ & \textit{relu} & \textit{relu} & \textit{relu} & \textit{relu} & \textit{relu} & - \\
batch size ($S$)  & 4 & 8 & 8 & 2 & 4 & - \\
number of iterations ($I_{\max}$) & 186750 & 287250 & 30750 & 63000 & 285000 & - \\
order of diffusion convolution & 2 & 2 & 2 & 2 & 2 & - & 2\\ 
    \bottomrule
    \end{tabular}
    \label{tab:para}
\end{table}

\textbf{KCN-Sage} We implemented a Pytorch version KCN following the code\footnote{https://github.com/tufts-ml/KCN} given by the original authors. The authors implemented several graph neural networks based kriging models, including GCN, GAT, and GraphSage. According to the experimental results, the KCN-Sage model achieves the best performance. So we compare SATCN with KCN-Sage. KCNs are originally proposed for kriging task under a fixed graph structure, we use Algorithm 1 to train them adpative to our task. As the same as \citep{appleby2020kriging}, a 4-layer GraphSage with hidden size [20, 10, 5, 3] is used. The values of 3 nearest neighbors are used as inputs for KCN-Sage.  
\bibliographystyle{ACM-Reference-Format}
\bibliography{acmart}